\let\origvec\vec
\documentclass[runningheads]{llncs}

\let\vec\origvec

\usepackage{amsmath}
\usepackage[utf8]{inputenc} % allow utf-8 input
\usepackage{hyperref}       % hyperlinks
\usepackage{url}            % simple URL typesetting
\usepackage{booktabs}       % professional-quality tables
\usepackage[ruled]{algorithm2e}
\usepackage{amsfonts}       % blackboard math symbols
\usepackage{nicefrac}       % compact symbols for 1/2, etc.
\usepackage{microtype}      % microtypography
\usepackage{color}
\usepackage[caption=false]{subfig}
\usepackage{graphicx}

\newcommand{\Acal}{\mathcal{A}}
\newcommand{\Scal}{\mathcal{S}}
\usepackage{graphicx}
\DeclareMathOperator*{\argmax}{arg\,max}
\begin{document}
	\title{Learning to Plan via a Multi-Step Policy Regression Method}
	\author{Stefan Wagner         \and
		Michael Janschek      \and
		Tobias Uelwer		  \and
		Stefan Harmeling}
	\authorrunning{S. Wagner et al.}
	\institute{Department of Computer Science, Heinrich Heine University Düsseldorf, Germany
		\email{\{stefan.wagner, michael.janschek, tobias.uelwer, stefan.harmeling\}@hhu.de}\\}
	\maketitle              % typeset the header of the contribution
	\begin{abstract}
		We propose a new approach to increase inference performance in environments that require a specific sequence of actions in order to be solved. This is for example the case for maze environments where ideally an optimal path is determined. Instead of learning a policy for a single step, we want to learn a policy that can predict $n$ actions in advance. Our proposed method called policy horizon regression (PHR) uses knowledge of the environment sampled by A2C to learn an $n$ dimensional policy vector in a policy distillation setup which yields $n$ sequential actions per observation. We test our method on the MiniGrid and Pong environments and show drastic speedup during inference time by successfully predicting sequences of actions on a single observation.
	\end{abstract}
	\section{Introduction}
	In recent years, reinforcement learning has seen growing success due to the use of deep learning as in \cite{mnih-atari-2013}. The reinforcement learning field has been mostly split between two major subfields: model-free and model-based reinforcement learning. In the first case we have a reactionary agent which learns directly from sampled experience, while in the latter case a model of the environment is learned from which the agent samples trajectories. Especially in model-based reinforcement learning the term of planning has come to fruition. Either prior or during agent training a model of the environment is learned. The agent then plans by simulating trajectories by using some form of tree search and thus is able to select the best actions while using fewer samples.
	
	Recently, there have been investigations about how to plan with a model-free based approach. Guez et al. \cite{guez2019investigation} trained an agent with a regular neural network with an architecture that has not been modified to enable any special planning behavior. The authors achieve state of the art performance for combinatorial problems such as Sokoban. However, after $10^{9}$ steps the authors experiments show that a simpler CNN seems to be sufficient in order to learn several possible variations of Sokoban. Another area that deals with planning in reinforcement learning is hierachical reinfocerment learning (HRL) as in \cite{hrl}. The main idea of HRL is not only to predict primitive actions for every time step, but macro-actions called \emph{options}. An option represents a policy that contains a sequence of primitive actions that are executed until a termination point. Then the agent may choose another option. HRL can be adapted to regular reinforcement learning via semi-markov decision processes (SMDPs). Policy distillation is a straightforward way of learning multiple tasks within a single policy. Policy distillation defines a \emph{teacher policy} $T$ and a \emph{student policy} $S$ where the teacher policy $T$ is usually first trained in advance and then the student policy $S$ is trained to match the teacher policy via supervised learning. We combine model-free reinforcement learning, HRL and policy distillation to create a new method to speed up model inference.
	
	We take the recent advances in model-free reinforcement learning by Guez et al. \cite{guez2019investigation} and take inspiration from HRL to create a method that leverages the inherent planning capability and simplicity of model-free learning, while applying the notion of options that stems from HRL to predict $n$ actions for a single state in order to speed up model inference. That is, for a single state we want to predict the following $n$ actions without changing the function approximator's architecture much. While the general outline in HRL is to define options over an SMDP, we learn an MDP and settle for a specific scenario, where the agent predicts a fixed number of actions for a given state. We do this by first training the environment with an extended A2C  architecture that first learns a base policy via regular A2C \cite{mnih2016asynchronous} which will serve as teacher policy $T$. In a second training stage, in order to learn $n$ actions for a single input, we take successful trajectories of the trained agent and regress these teacher policies on to the extended A2C architecture, so that it learns to predict $n$ actions given a single observation. Our method can be seen as part of intra-option learning as in \cite{intraoption} where options are learned off-policy from experience. In this way, we can reduce the number of evaluations needed to solve the environment during inference time. We thus formulate our main contributions with this work:
	\newline
	\newline
	\textbf{Contributions}
	\begin{itemize}
		\item We achieve model inference speedup in a reinforcement learning setting by leveraging the general framework of policy distillation, adding to the many use cases such as neural network compression and multi-task learning.
		\item We achieve a substantial inference speedup, as the prediction of the $n$ dimensional action vector is much more efficient than evaluating the model $n$ times.
		\item We show empirically that the inference speedup is due to the agent completing the environments faster than its non-PHR counterpart. Thus increasing the productivity of the agent.
		\item With our flexible and simple approach, we especially see a benefit in problems where the agent has limited resources during inference time or where the agents productivity should be boosted.
	\end{itemize}
	We propose our method as a viable option for \emph{optimal path finding}. Popular path-finding algorithms such as A* in \cite{hart1968formal} are able to find optimal paths given a start-point and an endpoint.  However it can be challenging to find a good heuristic that works for a given environment. Deep learning circumvents the need for a heuristic by learning directly from data.
	
	We demonstrate this by training PHR on two MiniGrid environments. In the first set of experiments the agent has to find the goal in a multi room grid and a stochastic grid that changes after every episode with the only reward being the end goal. We also train PHR on the Pong-Deterministic-v4 environment, showing that reactive environments which do not follow a grid structure can be enhanced with PHR as well. Overall, we achieve twice the inference performance while learning an optimal path i.e. sequences of actions while retaining policy quality.
	
	\section{Related Work}
	Sequencing actions to leverage different time scales has been used in reinforcement learning to improve learning speed. Schoknecht et al. \cite{schoknecht2002speeding} improve learning speed by defining multi-step actions on different time scales.
	The authors argue that between important actions (decisions) there may exist more superfluous actions that have to be executed repeatedly and should therefore only be learned as a single decision. This approach has also been extended to deep learning by Lakshminarayanan et al. \cite{dynamicar} and been formalized by Lee et al. \cite{leemultfreq}. Our approach focuses on sequencing actions as an abstraction over actions rather than for different time scales. By sequencing $n$ actions for a single state we reduce the number of model evaluations, thus speeding up inference time.
	
	Efroni et al. \cite{1pi} look at the problem of policy improvement with multi-step actions. The algorithm $h$-PI tries to find the choice of actions that maximizes the joint cumulativue expected reward given a sequence of actions within a defined horizon $h$ starting from a given state. De Asis et al. \cite{multirl} define a hyperparameter $\sigma$ to perform policy evaluation seamlessly over multiple steps.
	
	Panov et al. \cite{panov2018grid} used a CNN for path finding in conjunction with a custom reward function in settings that are challenging for A* with moderate results.
	
	Contrary to the work mentioned above, our approach focuses on speeding up model inference in real word applications while maintaining policy quality.
	Our approach is inspired by \emph{policy distillation} as in \cite{distill} and more generally specified in \cite{distilloverview}. Generally, the focus in knowledge distillation is to transfer knowledge to smaller networks. We seek to achieve inference speedup in a reinforcement learning setting by leveraging the general framework of policy distillation. In this way the architecture is only affected minimally and efficiency can be improved by either accelerating workflow or reducing the power target. Policy distillation defines a \emph{teacher policy} $T$ and a \emph{student policy} $S$ where the teacher policy $T$ is usually first trained in advance and then the student policy $S$ is trained to match the teacher policy via supervised learning. We learn a good policy via regular policy gradient ascent which serves as the teacher policy and then map a horizon of $n$ actions from the teacher policy to the student policy, which in this case is the function approximator which predicts $n$ actions for a single state.
	
	Improving inference performance for a CNN has been previously investigated by Ning et al. \cite{ning2019deep}. The authors use a collection of clustering techniques and similarity measures to compress the network and thus improve inference time. Our algorithm does not modify the network functionality, but leverages the reinforcement learning setting on an environment level.
	\section{Methods}
	In this section we describe our neural network architecture and our extension to the policy gradient called \emph{policy horizon regression} (PHR) which allows for an $n$ dimensional policy vector to be learned.
	\subsection{PHR Architecture}\label{architecture}
	PHR can be understood as an extension to any function approximator that requires planning its steps ahead of time. In this particular case we use A2C as the foundation for our training algorithm which includes the baseline loss and entropy regularization. We then extend the policy gradient in order to predict $n$ additional agent moves where $H=\left\{1, 2, \ldots , n\right\}$ is the \emph{policy horizon}. We say our extended architecture predicts an $n$ dimensional \emph{policy vector}.
	
	\subsubsection{Policy Vector.} A policy $\pi$ represents a probability distribution that assigns  probabilities $\pi(a|s)$ for
	each state $s\in \Scal$ over actions
	$a\in\Acal$, i.e., $\sum_a \pi(a|s)=1$ for all $s\in\Scal$.  We also
	write $\pi(\cdot | s)$ to denote the probability distribution over states for
	a fixed state $s\in\Scal$. Moreover, we denote $\pi^{\theta}$ as a policy determined by the network parameters $\theta$. Let $\pi_1^{\theta}$, \ldots, $\pi_n^{\theta}$ be a sequence of such policies, then we
	call the vector of these policies
	\begin{align}
		\vec{\pi}^{\theta} = [\pi_1^{\theta}, \ldots, \pi_n^{\theta}]^T
	\end{align}
	a \emph{policy vector}.  Note that for a given state $s\in\Scal$
	the policy vector $\vec{\pi}^{\theta}$ defines $n$ probability distributions $\pi_1^{\theta}, \ldots, \pi_n^{\theta}$ over the action space that are determined by the network parameters $\theta$.
	\begin{figure}[ht]
		\centering
		{\includegraphics[scale=0.4]{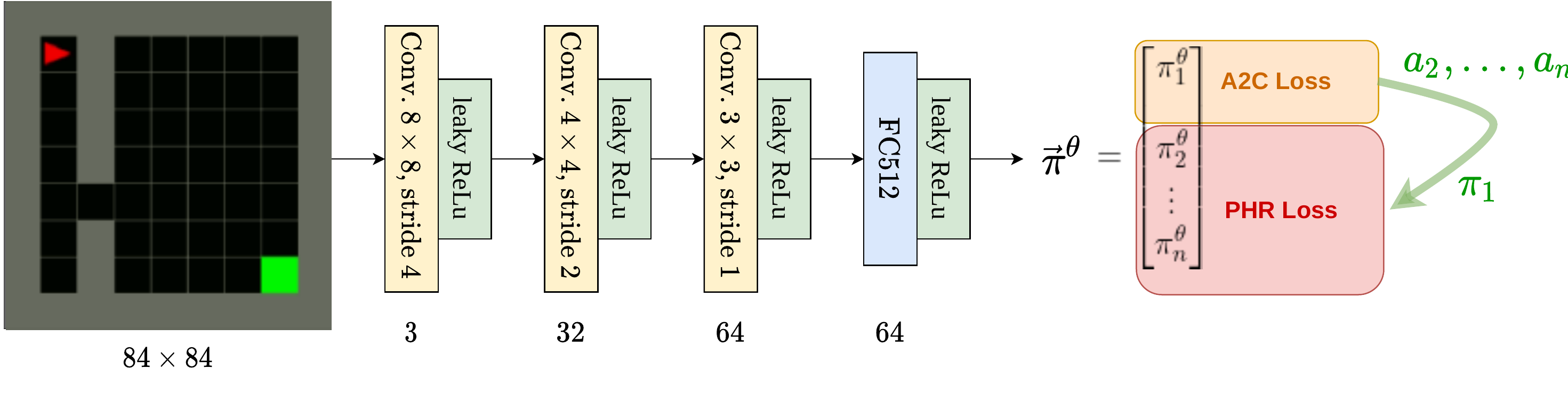}}
		\caption{Instead of predicting a single policy we predict a policy vector $\vec{\pi}^{\theta}$ by mapping $n$ actions or policies from $\pi_1^{\theta}$ to the policy vector.}
		\label{fig:fanope}
	\end{figure}
	Fig. \ref{fig:fanope} depicts the          PHR architecture: for a single state $s \in S$ we seek to predict a vector of \emph{n policies}. We use $\pi_1^{\theta}$ as the target policy to sample $n$ actions from it, then PHR maps these $n$ actions to the \emph{policy vector} $\vec{\pi}^{\theta}$. Alternatively, the same can be done between $n$ probability distributions of $\pi_1^{\theta}$ and of the \emph{policy vector}.  Note that the policy vector is computed only at the last layer (FC512), while the rest of the network is not any different from the usual setup to learn a single policy. Essentially, the network learns to predict the correct sequence of actions for a given input.  Extension of the convolutional layers are not necessary as the learned representations are already powerful enough to learn the proposed environments with PHR successfully. Overall, extending only the last layer creates minimal overhead.
	
	\subsection{Policy Horizon Regression (PHR)} \label{phr}
	\subsubsection{Distances between q-values.}
	Let $q^T=[q^{T}_1, \ldots, q^{T}_m]$ and $q^S=[q^S_1, \ldots, q^S_m]$ be two vectors describing the q-values corresponding to each possible action for the agent. In this case the former is the teacher policy and the latter the student policy. Then the distance between the two vectors can be measured with the squared loss $d_2$ or the KL-divergence $d_{KL}$ (which is strictly speaking not a distance since not symmetric),
	\begin{align}
		d_2(q^S,q^T) &= \sum_i |q^S_i-q^T_i|^2 & d_{KL}(q^S,q^T) = \sum_{i} \sigma(q^S_i) \log\frac{\sigma(q^S_i)}{\sigma(q^T_i)}
	\end{align}
	where $\sigma$ is the softmax function. These two distance measures are considered in our experiments, however, others are possible too.
	\subsubsection{Mapping actions directly to policy vector $\vec{\pi}^{\theta}$.} Let $a^{*} = \argmax q^T$ be the index of the maximum value over the discrete values of $q^T$, i.e., the best action. Then the cross entropy loss between $q^S$ and the best action $a^{*}$ reads as follows,
	\begin{align}
		L_{CE}(q^S,a^{*}) = \log q^S_{a^{*}}.
		\label{eq:ce}
	\end{align}
	Compared to the distance measures, this approach seeks to map the teacher policy onto the student policy directly. Overall, $d_2$ and $d_{KL}$ should yield a softer regression of the teachers actions, while $L_{CE}$ is a logistic regression. We consider all three approaches as methods to map our base policy $\pi_{1}^{\theta}$ onto the policy vector $\vec{\pi}^{\theta}$. In the following, we also refer to the teacher policy as the base policy $\pi_{1}^{\theta} = q^T$  and to the student policy as the policy vector $\vec{\pi}^{\theta} = q^S$.
	\begin{figure}[ht]
		\centering
		\includegraphics[scale=0.34]{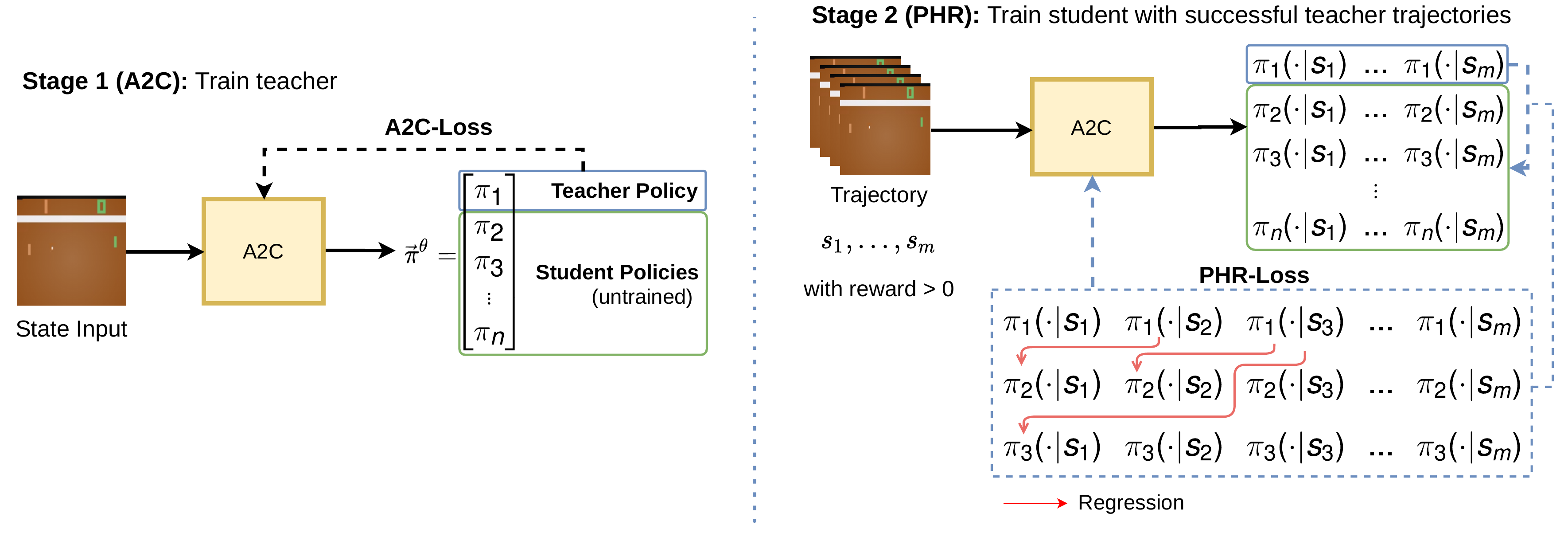}
		\caption{Diagram depicting PHR training process. (i) First, we train the teacher policy with regular A2C. (ii) In the second stage we evaluate successful trajectories from the teacher with reward $r_{m-1} > 0$ and regress the base policy $\pi_1^{\theta}$ onto the policy vector $\vec{\pi}^{\theta}$.}
		\label{fig:phr_train}
	\end{figure}
	
	\subsubsection{Learning policy vector $\vec{\pi}^{\theta}$ off-policy.}
	We learn the policy vector off-policy in two stages as in Fig. \ref{fig:phr_train}: (i) First the environment is learned fully with A2C via regular policy gradient ascent defined as
	\begin{align}
		J_{\text{PG}}(\theta) = \mathbb{E}_{\pi_{1}}\left[ \log \pi_{1}^{\theta}(s,a) \ Q^\theta(s,a)\right]
	\end{align}
	with corresponding gradient
	\begin{align}
		\nabla_{\theta} J_{\text{PG}}(\theta) &= \mathbb{E}_{\pi_1}\left[\nabla_\theta \log \pi_{1}^{\theta}(s,a) \ Q^\theta(s,a)\right].
		\label{eq:a2c_grad}
	\end{align}
	Note that the policy vector in our extended A2C architecture predicts $n$ policies. At this stage we have only learned the first policy $\pi_1^{\theta}$ which will serve as our \emph{teacher policy} to learn the rest of the policy vector.
	
	(ii) In the second stage we sample successful trajectories from our teacher policy $\pi_1^{\theta}$ (the so-called experience),
	\begin{align}
		D = \left\{ (s_1, \pi_1(\cdot, s_1), \ldots, s_m, \pi_1(\cdot, s_m)), \ldots\right\}
	\end{align}
	which will be regressed onto the policy vector, where $s_m$ is a terminal state with reward $r_{m-1} > 0$.
	We then take out sub-sequences $B_n$ of length $n$ from $D$
	\begin{align}
		B_n &= \left\{ (s_t, \pi_1(\cdot, s_t),  \ldots, s_{t+n-1},\pi_1(\cdot, s_{t+n-1})),
		\ldots\right\} & \text{with } 1\le t \le m -n+1 .
	\end{align} 
	We minimize the squared distance between the teacher policies $\pi_1^{\theta'}(\cdot | s_i)$ and the set of student policies $\pi_i^{\theta}(\cdot | s_t)$ of the sub-sequence
	\begin{align}\label{eq:5}
		\sum_{i=2}^n \left(\pi_{i}^{\theta}(\cdot|s_t)-\pi_1^{\theta'}(\cdot|s_i)\right)^2.
	\end{align}
	\indent Here, we use the squared distance, but any regression function can be used. Further note that $\theta$ is the parameter of the full policy vector
	$\vec{\pi}^{\theta}$.  For $\pi_1^{\theta'}$ the parameter is held fixed (denoted by $\theta'$). So PHR implicitly defines a \emph{semi-gradient} update method. Note, that PHR uses the rewards only to train the teacher policy $\pi_{1}^{\theta}$ via the A2C loss , but not for learning $\pi_2^{\theta}, \ldots, \pi_n^{\theta}$. In the second stage, the reward is only used to determine the best teacher trajectories.
	
	Finally, we have the PHR loss which is the expectation of Eq. \ref{eq:5} wrt. to all sub-sequences, i.e.,
	\begin{align}
		J_{\text{PHR}}(\theta, \theta') &= \mathbb{E}_{\mathcal{D}}\left[\sum_{i=2}^n \left(\pi_{i}^{\theta}(\cdot|s_t)-\pi_1^{\theta'}(\cdot|s_i)\right)^2\right]
		\label{eq:phr_loss}
	\end{align}
	where $(s_1, \ldots, s_n)$ are the random variables that are averaged out.
	The gradient for the PHR loss with respect to the parameters $\theta$ is:
	\begin{align} \label{eq:9}
		\nabla_\theta J_{\text{PHR}}(\theta, \theta') = \mathbb{E}_{\mathcal{D}}\left[\sum_{i=2}^n \nabla_\theta \pi_{i}^{\theta}(\cdot | s_t)  \left(\pi_{i}^{\theta}(\cdot |s_t)-\pi_1^{\theta'}(\cdot|s_i)\right)\right].
	\end{align}
	In other words, the policy vector $\vec{\pi}^{\theta}$ should learn to perform the same set of actions $a_2,\ldots,a_n$ just by looking at $s_t$ as $\pi_1^{\theta}$ would choose by looking at the full state sequence.
	Finally, this yields the following update rule,
	\begin{align}
		\nabla_\theta J(\theta) &= \lambda \nabla_\theta  J_{\text{PHR}}(\theta, \theta')
	\end{align}
	where $\lambda$ is a hyperparameter that adjusts the sensitivity of PHR and must be set depending on the chosen distance measure.  
	Overall, during the two stage process we combine the A2C gradient from Eq. \ref{eq:a2c_grad} with the PHR gradient which yields the full gradient for the parameter $\theta$ of the policy vector $\vec{\pi}^{\theta}$:
	\begin{align}
		\nabla_\theta J(\theta) &= \nabla_\theta J_{\text{PG}}(\theta) + \lambda \nabla_\theta  J_{\text{PHR}}(\theta, \theta')
	\end{align}
	The complete learning procedure is summarized in Algorithm \ref{alg:phr}.
	\SetAlFnt{\small}
	\SetAlCapFnt{\small}
	\SetAlCapNameFnt{\small}
	\begin{algorithm}[ht]
		Select horizon $n$, hyperparameter $\alpha\in \mathbb{N}$ and randomly initialize network with $\pi_1^{\theta}$ and  policy vector $\vec{\pi}^{\theta} = \left[\pi_1^{\theta}, \ldots, \pi_n^{\theta}\right]^T$.
		\BlankLine
		Train A2C agent fully as teacher with the A2C gradient
		to update $\pi_i^{\theta}$		
		\begin{align*}
			\nabla_{\theta} J_{\text{PG}}(\theta) &= \nabla_\theta \log \pi_{1}^{\theta}(s_t,a_t) \ Q^\theta(s_t, a_t)
		\end{align*} 
		
		\For{\text{episode}$ \ =1,\ldots, K$}{
			Sample \emph{experience} $(s_1, \pi_1(\cdot | s_1), \ldots, s_m, \pi_1(\cdot | s_m))$ from teacher policy $\pi_1^{\theta}$ where reward $r_{m-1} > 0$
			\BlankLine
			\For{$t =1, \ldots, m-n+1$}{
				\If{$t \mod \alpha == 0$}{
					Take out sub-sequence of $n$ states		
					$(s_t, \ldots, s_{t+n-1})$ from \emph{experience}
					
					\BlankLine
					
					Calculate PHR gradient to update policy vector $\vec{\pi}^{\theta}$
					\begin{align*} \label{eq:9}
						\nabla_\theta J_{\text{PHR}}(\theta, \theta') = \sum_{i=2}^n \sum_{j=1}^{A} \nabla_\theta \pi_{i}^{\theta}(a_{j} | s_t)	\left(\pi_{i}^{\theta}(a_{j}|s_t)-\pi_1^{\theta'}(a_{j}|s_{t+i-1})\right)
					\end{align*}
					\BlankLine
					Update model with PHR gradient
					\begin{align*}
						\nabla_\theta J(\theta) &= \lambda \nabla_\theta	J_{\text{PHR}}(\theta, \theta')
					\end{align*}
				}
			}
			
		}
		\caption{{\bf Policy Horizon Regression Algorithm} \label{alg:phr}}
	\end{algorithm}
	
	\subsubsection{Implementation Details}
	(i) Depending on the environment regressing every sub-sequence from Eq. \ref{eq:phr_loss} may lead to bad performance. Therefore, we define a hyperparameter $\alpha\in \mathbb{N}$ such that only every $\alpha$-th sub-sequence is used for regression. (ii) For $L_{CE}$ we sample the index of the best actions from the teacher policies as in Eq. \ref{eq:ce} and calculate the cross entropy loss between these actions and the student policies, i.e., $a^{*} = \argmax \pi_1(\cdot | s_t)$ and $L_{CE} = \log \pi_i(a^* | s_t) $.
	
	\section{Results}
	In the following experiments we investigate two contributions of PHR: (i) Is PHR able to increase inference speed in a meaningful way? (ii) Does the raw performance of the trained  agent scale according to the inference speedup? For this purpose we test PHR agents with horizons $n\in \{1, 4, 8, 16\}$ on two sets of environments, gym-minigrid and Pong from the ALE emulator \cite{ale}. To test (i) we measure the wall-clock time every agent needs to complete 100,000 steps. For (ii) we measure the amount of reward the agent can collect in one second.
	
	\subsection{Experiments}
	\begin{figure}[ht]
		\centering
		\subfloat{
			\includegraphics[width=0.25\linewidth]{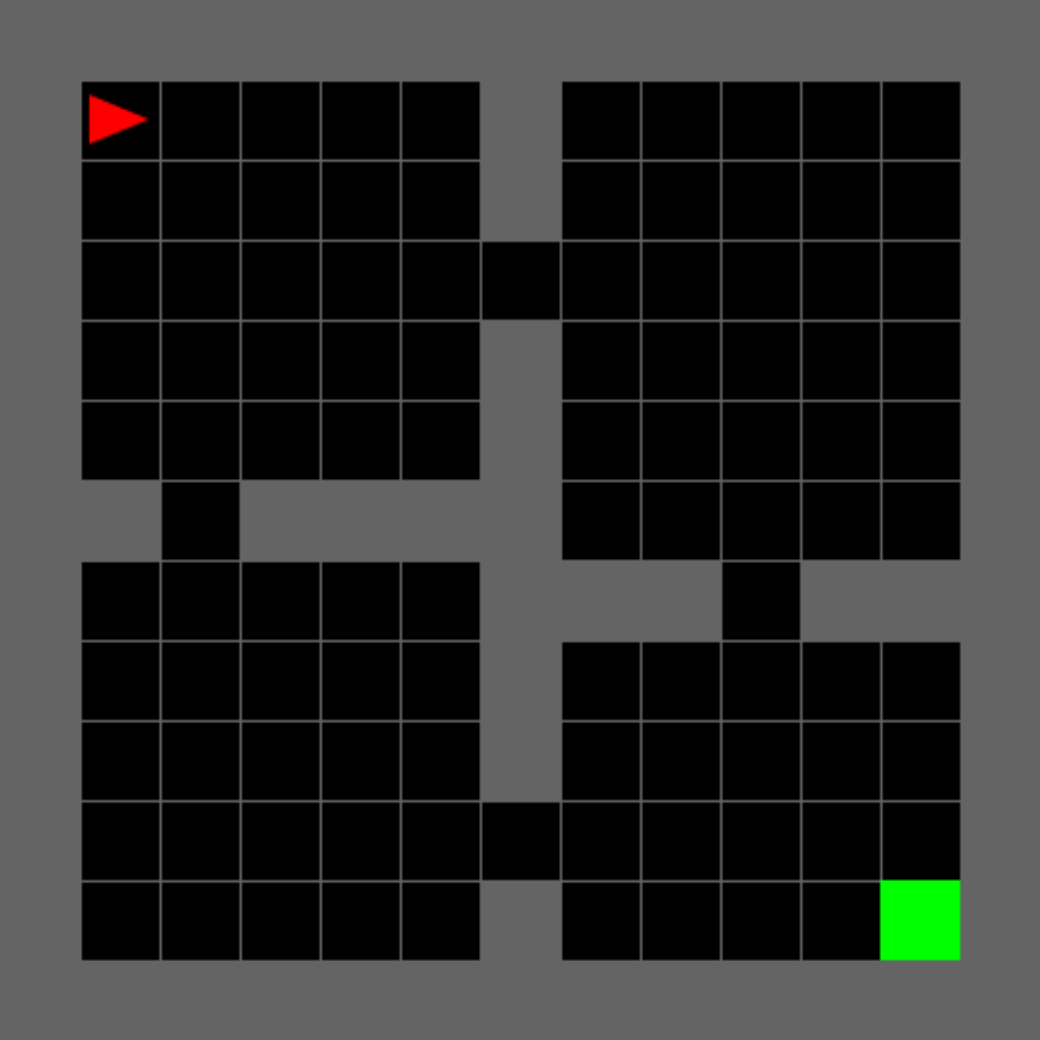}
		}%
		\hfil
		\subfloat{
			\includegraphics[width=0.25\linewidth]{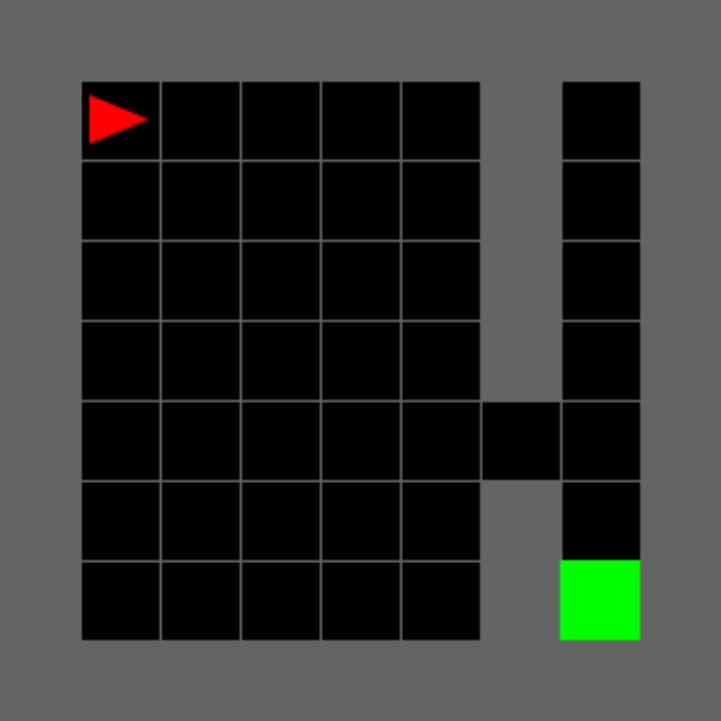}
		}%
		\hfil
		\subfloat{
			\includegraphics[width=0.25\linewidth]{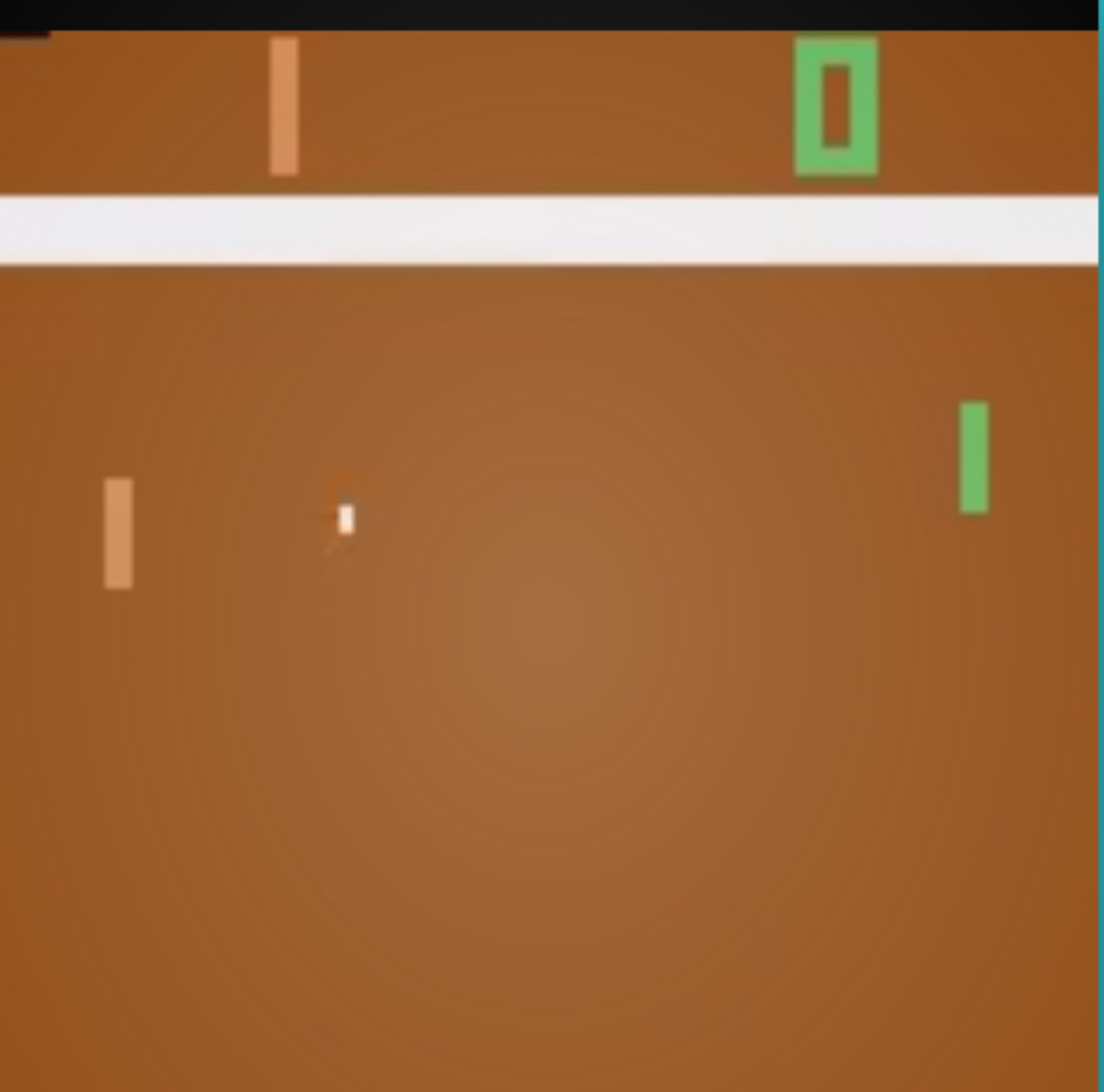}
		}%
		\caption{Environments used in this work. First we test on a deterministic grid environment (left) MultiRoom with $4$ rooms. The second environment (middle) Crossing is stochastic in the sense that the crossing changes position after every episode. Finally (right), we test on the Pong environment. }
		\label{crossing}
	\end{figure}
	The Python package gym-minigrid \cite{gym_minigrid} is a collection of gridworld environments. We chose the environments depicted in Fig. \ref{crossing} where the goal is to reach the green square. Only here the agent receives a reward making both environments very sparse. Changing direction and moving forward are two separate sets of actions. In addition, we chose the Pong-Deterministic-v4 environment from ALE \cite{ale}. The environment has $6$ different actions to interact with the players paddle. A reward signal of $1$ or $-1$ is given depending on which player scores a point. The episode ends when one player reaches the score of $21$. 
	
	\subsubsection{MultiRoom.} We adapted the MultiRoom environment to be the same as the environment in \cite{sutton1999between}, also used by \cite{schoknecht2002speeding}. It is a $13\times13$ grid and has $4$ crossings. This environment requires the agent to learn a long path from start to finish, thus we test how precisely PHR can map a set of actions to the policy vector.	
	
	\subsubsection{Crossing.} The SimpleCrossingS9N1 environments changes the position of the crossing after every episode with uniform probability to any position on the grid except the start and goal state. It is a $9\times 9$ grid with 1 crossing. We test whether PHR is able to learn multiple paths in an environment that changes frequently.
	
	\subsubsection{Pong.} Finally, we test PHR on a reactionary environment where the agent needs to react to an opponents actions. For this we use the Pong-Deterministic-v4 environment. We analyze whether PHR is able to learn where an agent must perform precisely timed actions and react to an opposing agents behavior.
	
	\subsection{Setup}
	To test PHR, we define agents that evaluate the environment every $n$ steps. In other words, these agents perform $n$ actions from a single state evaluation before evaluating the state again. For example: an agent that evaluates \emph{every} state to generate a policy is denoted with $n = 1$.  An agent that evaluates every \emph{fourth} state to generate a policy is denoted with $n = 4$. Moreover, once an agent completes an episode it is allowed to evaluate the model, regardless of the current action in the policy vector. We test every environment for $n \in \{1, 4, 8, 16\}$ and perform 5 runs for every configuration. For the Crossing environment the seed for the randomization of the environment is different across all $5$ runs. 
	
	\subsubsection{Performance metrics.} We define two performance metrics for our experiments. We measure wall-clock time and score per second between different horizons $n$ for PHR. With wall-clock time we measure the overall inference speedup provided by evaluating the policy vector instead of the complete model. The score per second measures the reward the agents can gather in one second. This allows us to measure whether the agents raw performance also scales with inference speedup. In this way we can visualize the trade-off between faster model evaluation and the agents quicker task completion due to the sequencing of actions.
	
	\subsubsection{Baseline.} We define $n=1$ as the baseline. This is the agent that evaluates the model at every step, i.e., regular model inference with A2C. Ideally, the PHR agents $n\in \{4, 8, 16\}$ should be more efficient to evaluate and complete the environments faster while maintaining or increasing raw performance. 
	
	\subsection{Performance}
	\begin{figure}[ht]
		\centering
		\includegraphics[width=\linewidth]{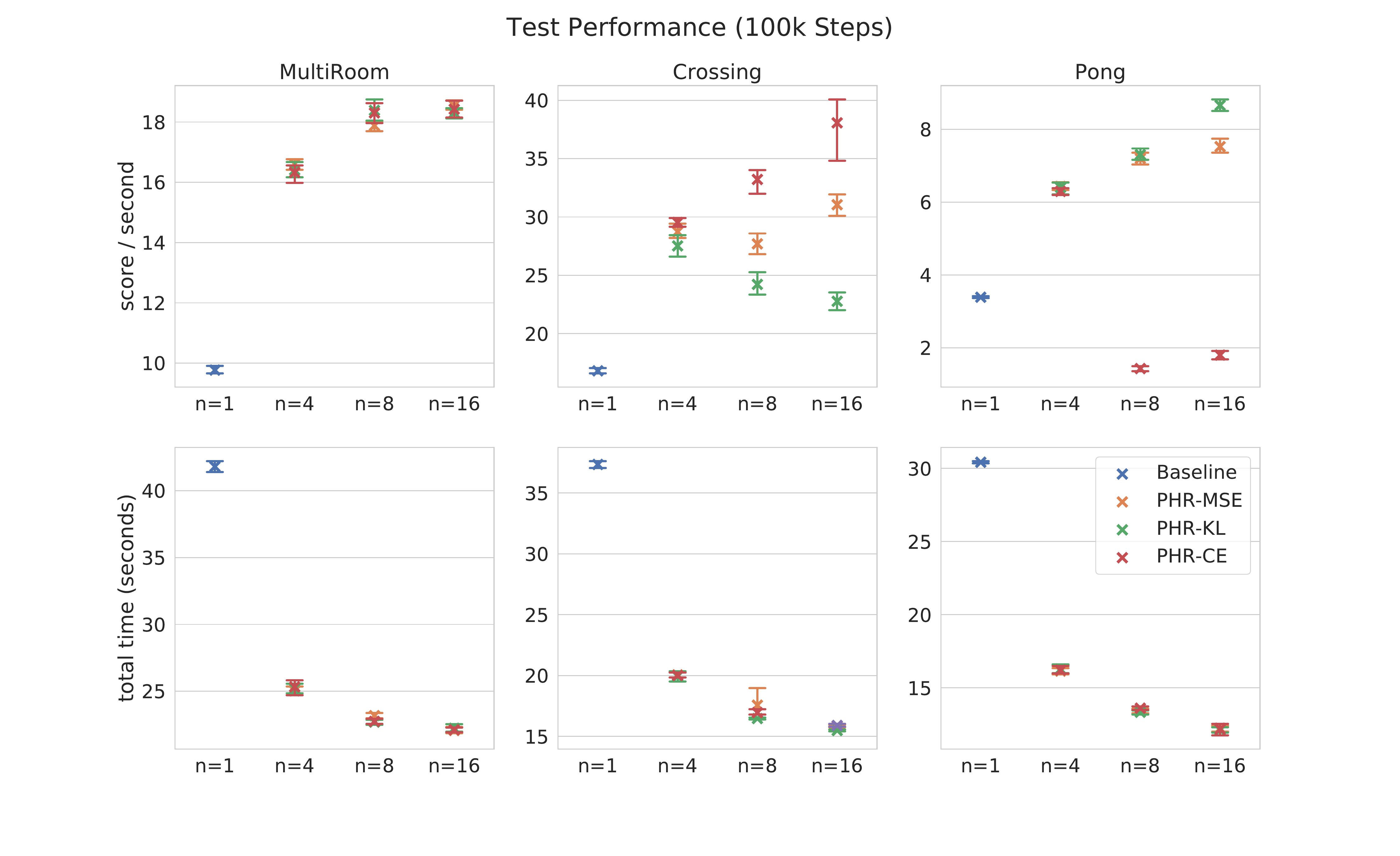}
		\caption{Score per second and total inference time ($100$k steps, 5 runs) for all $3$ environments. The top 3 plots show the achieved score per second (higher is better), while the lower $3$ plots show the time needed to complete $100$k steps (lower is better). PHR is generally able to reduce the wall-clock time  at least by half, thus doubling inference speed. Moreover, PHR is able to maintain policy quality thus achieving a higher throughput.} 
		\label{fig:results}
	\end{figure}
	
	Fig. \ref{fig:results} shows the wall-clock time of the different agents and their throughput in score per second for $100$k steps of inference averaged over $5$ runs. We generally see that PHR is able to provide at least double the inference speedup in all $3$ environments, effectively only needing at least half the time to complete $100$k steps, while scaling logarithmically through the different agents (lower part of Fig. \ref{fig:results}). Moreover, as policy quality is maintained, the agent is able to increase its throughput by the same factor (upper part of Fig. \ref{fig:results}). This means that not only is evaluating the policy vector more efficient, but predicting $n$ actions for a single observation also effectively increases the throughput i.e the productivity of the agent. Due to the low tolerance of error in Crossing,  cross entropy leads to vastly better performance as it maps actions directly compared to the distance measures. Pong on the other hand benefits from a softer regression as this produces less jittery behavior which is counterproductive in reactive settings. 
	
	\subsection{Learned Path}
	\begin{figure}[ht]
		\centering
		\includegraphics[scale=0.65]{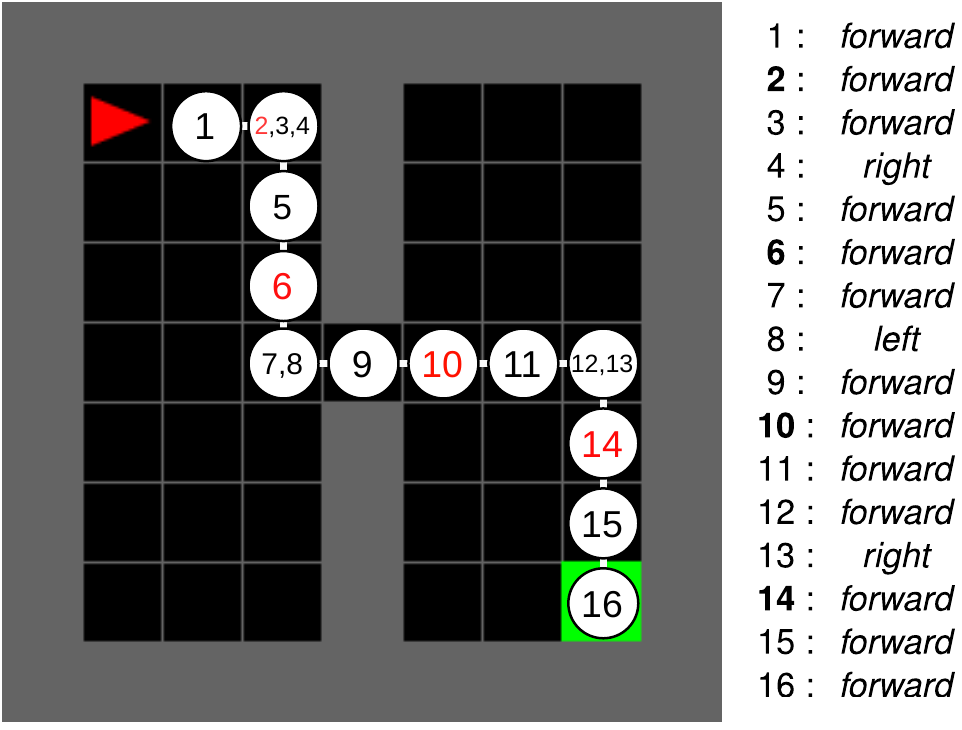}
		\caption{Learned path for one configuration of the Crossing environment with two rooms. Crucial actions are taken at intermediate policies, thus showing that PHR has learned a path. Numbers $1$ to $16$ represent  actions performed to land in each state. {\color{red}Red} (\textbf{Bold}) numbers represent a state evaluation, which happens every $4$ actions.}
		\label{fig:learnedpath}
	\end{figure}
	%\vspace{-4pt}
	We analyzed the paths that are actually learned by the agent. Fig. \ref{fig:learnedpath} shows the learned path of an agent evaluating $n=4$, i.e., evaluating a state every 4 actions.
	We see that the agent is able to learn the most important decisions such as turning towards the crossing in actions $4, 8$ and turning towards the goal in action $13$. Moreover, these actions are determined from intermediate policies that have been learned by PHR.
	
	\section{Discussion}
	\subsubsection{Limitations.} A drawback to our method is that we require the tasks to be sequenceable and not highly stochastic. As the agent is committed to $n$ steps when it evaluates the policy vector, all stochastic behavior must be observable to the agent during the first stage of training the teacher policy.
	However, we show with the Crossing and Pong environments that PHR is well able to handle stochasticity and reactive settings. Thus, once the teacher policy is successfully learned, learning the policy vector is straightforward and remains performant. 
	
	\subsubsection{Conclusion.}Overall, we succeeded in training an agent to predict $n$ actions given only a single state. We showed on one hand that an optimal path can be learned with PHR in the minigrid environments. On the other hand we showed that PHR is well capable of being used in a reactive environment that is not sequenced as an optimal path in a grid, but as an optimal set of actions that have to be executed precisely. With PHR we drastically reduce the computational cost at inference time as it takes less time to evaluate the policy head with $n$ actions than to evaluate the model $n$ times. Furthermore, PHR sequences actions reliably such that it is able to complete the environments faster, providing an even greater inference speedup. This opens PHR up to easy implementation in real-world applications where limited computing resources are of concern.% or where productivity should be maximized.

	\bibliographystyle{splncs04}
	\bibliography{paper}

\begin{thebibliography}{10}
\providecommand{\url}[1]{\texttt{#1}}
\providecommand{\urlprefix}{URL }
\providecommand{\doi}[1]{https://doi.org/#1}

\bibitem{hrl}
Barto, A., Mahadevan, S.: Recent advances in hierarchical reinforcement
  learning. Discrete Event Dynamic Systems: Theory and Applications
  \textbf{13} (12 2002). \doi{10.1023/A:1025696116075}

\bibitem{ale}
{Bellemare}, M.G., {Naddaf}, Y., {Veness}, J., {Bowling}, M.: {The Arcade
  Learning Environment: An Evaluation Platform for General Agents}. arXiv
  e-prints arXiv:1207.4708 (Jul 2012)

\bibitem{gym_minigrid}
Chevalier-Boisvert, M., Willems, L., Pal, S.: Minimalistic gridworld
  environment for openai gym. \url{https://github.com/maximecb/gym-minigrid}
  (2018)

\bibitem{distilloverview}
{Czarnecki}, W.M., {Pascanu}, R., {Osindero}, S., {Jayakumar}, S.M.,
  {Swirszcz}, G., {Jaderberg}, M.: {Distilling Policy Distillation}. arXiv
  e-prints arXiv:1902.02186 (Feb 2019)

\bibitem{multirl}
{De Asis}, K., {Hernandez-Garcia}, J.F., {Zacharias Holland}, G., {Sutton},
  R.S.: {Multi-step Reinforcement Learning: A Unifying Algorithm}. arXiv
  e-prints arXiv:1703.01327 (Mar 2017)

\bibitem{1pi}
{Efroni}, Y., {Dalal}, G., {Scherrer}, B., {Mannor}, S.: {Beyond the One Step
  Greedy Approach in Reinforcement Learning}. arXiv e-prints arXiv:1802.03654
  (Feb 2018)

\bibitem{guez2019investigation}
Guez, A., Mirza, M., Gregor, K., Kabra, R., Racanière, S., Weber, T., Raposo,
  D., Santoro, A., Orseau, L., Eccles, T., Wayne, G., Silver, D., Lillicrap,
  T.: An investigation of model-free planning (2019)

\bibitem{hart1968formal}
Hart, P.E., Nilsson, N.J., Raphael, B.: A formal basis for the heuristic
  determination of minimum cost paths. IEEE transactions on Systems Science and
  Cybernetics  \textbf{4}(2),  100--107 (1968)

\bibitem{dynamicar}
Lakshminarayanan, A.S., Sharma, S., Ravindran, B.: Dynamic action repetition
  for deep reinforcement learning. In: Proceedings of the Thirty-First AAAI
  Conference on Artificial Intelligence. p. 2133–2139. AAAI'17, AAAI Press
  (2017)

\bibitem{leemultfreq}
Lee, J., Lee, B.J., Kim, K.E.: Reinforcement learning for control with multiple
  frequencies. In: Larochelle, H., Ranzato, M., Hadsell, R., Balcan, M.F., Lin,
  H. (eds.) Advances in Neural Information Processing Systems. vol.~33, pp.
  3254--3264. Curran Associates, Inc. (2020),
  \url{https://proceedings.neurips.cc/paper/2020/file/216f44e2d28d4e175a194492bde9148f-Paper.pdf}

\bibitem{mnih2016asynchronous}
Mnih, V., Badia, A.P., Mirza, M., Graves, A., Lillicrap, T., Harley, T.,
  Silver, D., Kavukcuoglu, K.: Asynchronous methods for deep reinforcement
  learning. In: International conference on machine learning. pp. 1928--1937.
  PMLR (2016)

\bibitem{mnih-atari-2013}
Mnih, V., Kavukcuoglu, K., Silver, D., Graves, A., Antonoglou, I., Wierstra,
  D., Riedmiller, M.: Playing atari with deep reinforcement learning. In: NIPS
  Deep Learning Workshop (2013)

\bibitem{ning2019deep}
Ning, L., Shen, X.: Deep reuse: streamline cnn inference on the fly via
  coarse-grained computation reuse. In: Proceedings of the ACM International
  Conference on Supercomputing. pp. 438--448 (2019)

\bibitem{panov2018grid}
Panov, A.I., Yakovlev, K.S., Suvorov, R.: Grid path planning with deep
  reinforcement learning: Preliminary results. Procedia computer science
  \textbf{123},  347--353 (2018)

\bibitem{distill}
{Rusu}, A.A., {Gomez Colmenarejo}, S., {Gulcehre}, C., {Desjardins}, G.,
  {Kirkpatrick}, J., {Pascanu}, R., {Mnih}, V., {Kavukcuoglu}, K., {Hadsell},
  R.: {Policy Distillation}. arXiv e-prints arXiv:1511.06295 (Nov 2015)

\bibitem{schoknecht2002speeding}
Schoknecht, R., Riedmiller, M.: Speeding-up reinforcement learning with
  multi-step actions. In: International Conference on Artificial Neural
  Networks. pp. 813--818. Springer (2002)

\bibitem{intraoption}
Sutton, R., Precup, D., Singh, S.: Intra-option learning about temporally
  abstract actions. pp. 556--564 (01 1998)

\bibitem{sutton1999between}
Sutton, R.S., Precup, D., Singh, S.: Between mdps and semi-mdps: A framework
  for temporal abstraction in reinforcement learning. Artificial intelligence
  \textbf{112}(1-2),  181--211 (1999)

\end{thebibliography}
	
\end{document}